\documentclass[pdflatex,bst/sn-mathphys-num]{sn-jnl}% Math and Physical Sciences Numbered Reference Style
\usepackage{graphicx}%
\usepackage{multirow}%
\usepackage{amsmath,amssymb,amsfonts}%
\usepackage{amsthm}%
\usepackage{mathrsfs}%
\usepackage[title]{appendix}%
\usepackage{xcolor}%
\usepackage{textcomp}%
\usepackage{manyfoot}%
\usepackage{booktabs}%
\usepackage{algorithm}%
\usepackage{algorithmicx}%
\usepackage{algpseudocode}%
\usepackage{listings}%
\usepackage{adjustbox}
\usepackage[numbers]{natbib}
\usepackage{subcaption} % Add this in your preamble
\usepackage{adjustbox}  % Helps with resizing content inside tables
\usepackage{graphicx}
\usepackage{booktabs}
\usepackage{caption}
\usepackage{wrapfig}

\usepackage{todonotes}

\theoremstyle{thmstyleone}%
\theoremstyle{thmstyletwo}%

\theoremstyle{thmstylethree}%

\begin{document}

\title[Article Title]{Beyond Fixed Topologies: Unregistered Training and Comprehensive Evaluation Metrics for 3D Talking Heads}


\author[1,*]{\fnm{Federico} \sur{Nocentini}}%\email{federico.nocentini@unifi.it}
\equalcont{These authors contributed equally to this work.}

\author[2, *]{\fnm{Thomas} \sur{Besnier}}%\email{thomas.besnier@univ-lille.fr}
\equalcont{These authors contributed equally to this work.}

\author[4]{\fnm{Claudio} \sur{Ferrari}}%\email{claudio.ferrari@unisi.it}

\author[5]{\fnm{Sylvain} \sur{Arguillere}}%\email{sylvain.arguillere@univ-lille.fr}

\author[2,3]{\fnm{Mohamed} \sur{Daoudi}}%\email{mohamed.daoudi@imt-nord-europe.fr}

\author[1]{\fnm{Stefano} \sur{Berretti}}%\email{stefano.berretti@unifi.it}

\affil[1]{MICC, University of Florence}
\affil[2]{CRIStAL, University of Lille, CNRS, Centrale Lille}
\affil[3]{IMT Nord Europe, University of Lille}
\affil[4]{Dept. of Information Engineering and Mathematics, University of Siena}
\affil[5]{Laboratoire Paul Painlevé, University of Lille, CNRS}
%\affil[1]{Media Integration and Communication Center (MICC), University of Florence, \city{Florence}, \country{Italy}}

%\affil[2]{Univ. Lille, CNRS, Centrale Lille, UMR 9189 CRIStAL, F-59000, \orgaddress{\city{Lille}, \country{France}}}

%\affil[3]{IMT Nord Europe, Institut Mines-Télécom, Univ. Lille, Centre for Digital Systems, F-59000, \city{Lille}, \country{France}}

%\affil[4]{Dept. of Information Engineering and Mathematics, Univ. of Siena, \city{Siena}, \country{Italy}}

%\affil[5]{Univ. Lille, CNRS, UMR 8524 Laboratoire Paul Painlevé, F-59000, \orgaddress{\city{Lille}, \country{France}}}

%\affil[3]{\orgdiv{Department}, \orgname{Organization}, \orgaddress{\street{Street}, \city{City}, \postcode{610101}, \state{State}, \country{Country}}}


\abstract{Generating speech-driven 3D talking heads presents numerous challenges; among those is dealing with varying mesh topologies where no point-wise correspondence exists across the meshes the model can animate. 
While previous literature works assume fixed mesh structures,  in this work we present the first framework capable of animating 3D faces in arbitrary topologies, including real scanned data. Our approach leverages heat diffusion to predict features that are robust to the mesh topology. We explore two training settings: a registered one, in which meshes in a training sequences share a fixed topology but any mesh can be animated at test time, and an fully unregistered one, which allows effective training with varying mesh structures. Additionally, we highlight the limitations of current evaluation metrics and propose new metrics for better lip-syncing evaluation. An extensive evaluation shows our approach performs favorably compared to fixed topology techniques, setting a new benchmark by offering a versatile and high-fidelity solution for 3D talking heads where the topology constraint is dropped. The code along with the pre-trained model, is available \href{https://github.com/miccunifi/ScanTalk}{here}.}

\maketitle
%\footnote{* These authors contributed equally to this work.}

% \begin{figure}[ht]
%     \vspace{-1.5cm}
%     \centering
%     \includegraphics[width=0.7\linewidth]{img/teaser_new.png}
%     \caption{This paper showcases a framework capable of animating \textbf{any} 3D face mesh driven by speech. In particular, it is robust enough to learn from multiple unrelated datasets using a single model, accommodating both registered and unregistered meshes.}
%     \label{fig:idea}
%     \vspace{-1cm}
% \end{figure}
\vspace{4cm}

\section{Introduction}

\begin{wrapfigure}{r}{0.5\linewidth}
    \vspace{-1cm}
    \centering
    \includegraphics[width=\linewidth]{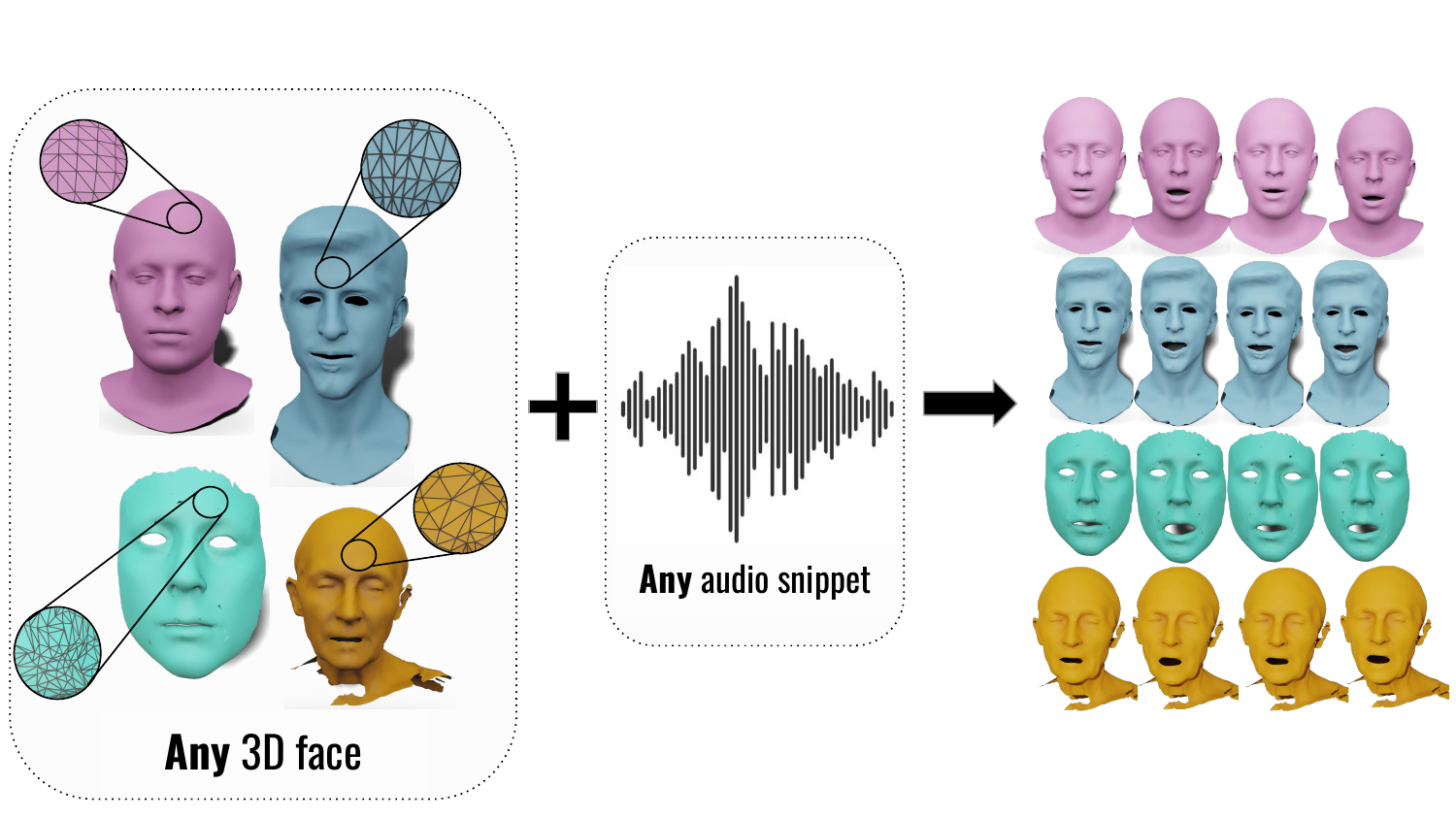}
    \caption{ScanTalk animates \textbf{any} 3D face mesh from speech, handling both registered and unregistered meshes across diverse datasets with a single model.}
    \label{fig:idea}
    \vspace{-0.6cm}
\end{wrapfigure}
3D talking heads refer to the task of animating a generic 3D face mesh so that it corresponds to a speech, given in the form of an audio signal. This task has received significant attention in recent years, thanks to the wide range of potential applications, across diverse fields, that it can open. This includes computer-generated imagery in movies, video games, virtual reality, and medical simulations. 
Nevertheless, it also faces unique challenges, primarily arising from the complexity of mapping the semantics of the spoken words contained in the audio to the intricacy of facial motions. 
A major challenge for 3D talking heads is that of generating convincing and audio-synchronized lip movements that faithfully replicate the spoken sentence. %This is the primary goal, and arguably the most challenging.
Several other aspects, though, contribute to making a generated animation realistic. For example, head movements, facial expressions or speech styles are all important factors, and numerous approaches were proposed to account for such diverse challenges~\cite{nocentini2024emovoca, peng2023emotalk, FaceDiffuser_Stan_MIG2023}. In all cases, the primary objective is that of learning a mapping between the audio input to non-rigid and time-dependent deformations of the face mesh that allow for its animation. 

In this regard, all state-of-the-art methods in this field simplify the problem by assuming that the meshes to be animated are aligned to a fixed topology, that is, all face meshes share a consistent number and arrangement of points. On one hand, this constraint allows for eluding most difficulties related to processing unordered point-clouds. With the meshes in vertex-wise correspondence and their structure known and consistent, one can, for instance, use effective graph operators~\cite{bouritsas_2019_neural3DMM, chebnet_2016, gong2019spiralnet++} or define specific losses to control localized face areas~\cite{Thambiraja_2023_ICCV_imitator}. On the other hand, it puts several constraints on the applicability. For example, existing approaches require training separate models for each specific topology, hindering its generalization power. In addition, if one wants to animate a new mesh, either captured with a real scanner or designed with some graphics software, it is first required to process the mesh to make it consistent with the topology used for training. This also adds a level of complexity in the data collection since newly captured mesh sequences need to be processed in order to align them to a common topology, which unveils new challenges mostly when non-rigid face deformations are involved~\cite{ferrari2021sparse, gilani2017dense}. 
In \cite{nocentini2024scantalk3dtalkingheads}, the training process is supervised by the vertex-wise correspondence in the data and the model is robust to generalize to different topologies at inference time. A step further is not assuming vertex-wise correspondence within sequences during the training process (unregistered sequences). In this scenario, training is not supervised by vertex-wise correspondence. This is especially interesting because it allows one to train on minimally processed data, enhancing even more the versatility of the method.

In addition to the aforementioned issues, another critical aspect is related to the evaluation protocols and metrics. Thoroughly assessing the performance of a 3D talking head method is complex, as it requires to simultaneously verify the extent to which: \textit{(i)} the overall shape of the animated face is preserved. There is in fact no guarantee that mouth motions won't negatively affect other face regions; \textit{(ii)} the accuracy of the lip motion and synchronization is maintained, both in terms of shape similarity, perceived realism and motion dynamics. We identified a clear shortcoming in the current literature in this regard. In most works, the primary metrics used for quantitative evaluation and comparison measure the discrepancy between ground-truth and generated animation in terms of per-frame geometric discrepancy of vertices~\cite{VOCA2019, Fan_Lin_Saito_Wang_Komura_faceformer_2022, peng2023selftalk, FaceDiffuser_Stan_MIG2023}. 
These are referred to as Lip-Vertex Error (LVE), Mean Vertex Error (MVE) and Face Dynamics Deviation (FDD). 
%LVE measures the average maximum Euclidean distance of the mouth vertices, while MVE considers the whole mesh. FDD, instead, focuses on the upper face part, and is intended to provide a measure of the animation diversity. 
We argue that these metrics do not provide a thorough picture of a method performance, and while these quantitative measures are widely used, they often fall short in providing a comprehensive evaluation. Suffice it to say, the evaluation of the motion dynamics and synchronization is largely ignored, and lower metrics do not necessarily imply a more realistic animation. A common workaround involves performing qualitative user studies, where people are asked to evaluate the quality of the generated samples. While useful to some extent, these studies lack a clear robust protocol, and usually involve a few users and examples, posing doubts on their reliability. Some papers raised similar concerns, and defined additional metrics to fill in such gap. For instance, in \cite{peng2023selftalk} the Lip Readability Percentage (LRP) is introduced. However, it is still based on per-frame lip vertex distances. Other examples are Dynamic Time Warping (DTW) used in Imitator \cite{Thambiraja_2023_ICCV_imitator} for assessing temporal coherence, or the Displacement Angle Discrepancy (DAE) proposed in \cite{landmarks_3D_Nocentini_2023}, which provides a measure of the angular difference between subsequent frames. These measures, and others, are a clear evidence that the current benchmarking miss a piece, and that a stable, consistent and comprehensive set of complementary error measures needs to be defined for proper evaluation across methods.

In response to such limitations, we introduce \textbf{ScanTalk}, the first framework capable of animating 3D faces regardless of their topology. ScanTalk is designed to be versatile, accommodating any mesh topology, including those from real 3D face scans. By maintaining fidelity and coherence across different topologies, ScanTalk ensures that facial animations remain realistic and expressive despite variations in the mesh structure. This adaptability is particularly crucial for speech-driven facial animations, where the dynamic nature of speech-related lip movements and associated facial changes demands a high-level of flexibility in the mesh topology. 
To address the challenges associated with unregistered meshes, we present new experiments that focus on handling unregistered point clouds and employing suitable loss functions for training. These experiments demonstrate ScanTalk's capability to effectively manage real-world scenarios, where mesh registration is impractical. In addition, we propose and extensively analyze a complementary set of error measures, with the aim of providing a complete view of the pros and cons of each, and showing how using multiple metrics can help for the sake of a comprehensive evaluation.

In summary, ScanTalk, illustrated in Figure~\ref{fig:idea}, represents a significant advancement in the field of speech-driven 3D face animation by overcoming the limitations of fixed-topology models and establishing new standards for evaluation and training. Our framework demonstrates broad applicability across different datasets and conditions. %potentially revolutionizing the way 3D facial animations are generated and assessed. 
Parts of the current work have been introduced in our previous paper~\cite{nocentini2024scantalk3dtalkingheads}. The present extension provides the following new contributions:
\begin{itemize}
    \item In~\cite{nocentini2024scantalk3dtalkingheads}, ScanTalk needs to be trained with homogeneous sequences: each frame of the sequence shares the topology of the first frame. Here, we modify the training scheme and define appropriate losses so that the model can be trained with inhomogeneous sequences: each frame of the sequence has an arbitrary topology. By dropping this constraint, ScanTalk can be trained with \textit{truly} unregistered meshes; 
    \item While the Chamfer distance is widely used as a loss for unsupervised static 3D reconstruction, we propose to use a dynamic extension of this loss for a complete unsupervised setting to learn speech-driven motion prediction;
    \item We expose and analyze the limitations of the current benchmark metrics for evaluating lip-sync and generation quality, prompting the introduction of new, comprehensive metrics. With these, we provide an extended and comprehensive evaluation of ScanTalk and state-of-the-art methods.
%\item \textcolor{red}{Others?}
\end{itemize}

\section{Related work}
\subsection{Animation of Unregistered 3D Face Meshes}
Animating a 3D face mesh to reproduce expressions or speech is a research topic that witnessed noticeable advancements only recently, thanks to the progresses achieved in the design of effective deep architectures for processing 3D data~\cite{bouritsas_2019_neural3DMM, lu2024fc, otberdout2022sparse, otberdout2023generating}. However, it is well known that deep models are data-hungry, which poses the problem of collecting sufficiently large 4D datasets. Moreover, all state-of-the-art methods based on deep learning require the face meshes to be in a fixed topology. 

In the realm of 3D face mesh acquisition for datasets collection, methodologies vary from extracting geometry from images or videos~\cite{yi2022_talkshow} or generating synthetic data~\cite{principi2023florence}, to employing specialized scanners in controlled environments~\cite{BIWI_2010, ferrari2023florence, COMA:ECCV18, wuu2022multiface, BU3DFE_2006}. 
While methods that follow the former paradigm are simpler and cost-effective, and typically result in meshes with known topology, they may not capture complete 3D information with the necessary fidelity. 
On the other hand, 3D scans, though rich in captured details of the face, often come unregistered. The varying mesh structure and nuisances resulting from the sensors, \textit{e.g.}, holes or noise, further complicate a direct animation. In response, a widely used approach to address this challenge is that of first registering the raw scans onto predefined topologies, typically by means of 3D Morphable Models~\cite{BIWI_2010, ferrari2021sparse, FLAME:SiggraphAsia2017, muralikrishnan_2023_BLISS, wuu2022multiface}. 

While some deep learning approaches have emerged to handle face scans with unknown structure through latent representations with robust encoders~\cite{sharp2021diffusion, Wiersma2022DeltaConv} built upon PointNet~\cite{Charles_PointNet_2017, qi2017pointnet++} or adaptive convolution strategies~\cite{dgcnn}, these methods still rely on a registered decoding strategy, which can smooth out important details. 
Recent advancements closer to our work include DiffusionNet~\cite{sharp2021diffusion} combined with neural Jacobian fields~\cite{NeuralJacobianField_2022} in Neural Face Rigging (NFR)~\cite{Qin_2023_NFR}, which facilitates animation transfer between sequences of unregistered face meshes. Our framework stands out by introducing a model that leverages audio data to animate unregistered meshes directly.

\subsection{3D Talking Heads}
Numerous methodologies have emerged to synchronize facial animations with speech. While significant progress has been made in 2D talking heads~\cite{EAMM_2022_2Dtalkinghead, Wang2023Emotional_2Dtalkinghead}, extending these approaches to 3D faces remains challenging. 
Nevertheless, generating 3D talking heads from audio has seen significant advancements, initially pioneered by~\cite{VOCA2019} and greatly improved through deep learning methods. Recent works leverage large datasets and deep learning strategies~\cite{VOCA2019, Fan_Lin_Saito_Wang_Komura_faceformer_2022, facexhubert, landmarks_3D_Nocentini_2023, nocentini2024emovoca, richard2021meshtalk, FaceDiffuser_Stan_MIG2023, Thambiraja_2023_ICCV_imitator, xing2023codetalker}. 
VOCA~\cite{VOCA2019} was among the first to develop a generalizable model for 3D talking heads but lacked expressiveness and convincing lip synchronization. MeshTalk~\cite{richard2021meshtalk} improved upon this with a larger registered dataset and a more expressive model. FaceFormer~\cite{Fan_Lin_Saito_Wang_Komura_faceformer_2022} utilized a transformer architecture and a pre-trained audio encoder, like Wav2vec2~\cite{wav2vec_2019}, with further improvements in FaceXHubert~\cite{facexhubert}, demonstrating enhanced cross-modality mappings from audio to facial motion. From a different perspective, Nocentini \textit{et al.}~\cite{landmarks_3D_Nocentini_2023} designed a coarse-to-fine approach named S2L-S2D, and showed that lip motion can be effectively modeled by learning to first animate facial landmarks, and then reconstructing the whole face shape. 
Recent models like CodeTalker~\cite{xing2023codetalker} and SelfTalk~\cite{peng2023selftalk} introduced new techniques to address over-smoothing and enhance lip motion expressiveness. Integrating emotions~\cite{danvevcek2023emotional, peng2023emotalk} and head poses~\cite{sun2023diffposetalk, SadTalker_2023} into models further pushes the boundaries, though the lack of registered data remains a significant challenge. In this regard, a recent work attempted to bridge this gap by proposing a data-driven approach to synthetically generate a large-scale dataset of expressive 3D talking heads, named EmoVOCA~\cite{nocentini2024emovoca}, which combines the VOCAset dataset~\cite{VOCA2019} with 4D expressive sequences from the Florence4D dataset~\cite{principi2023florence}. However, creating this dataset was made easy as both VOCAset and Florence4D contain meshes in the same topology. 
More recent approaches leverage denoising diffusion models for generating 3D talking heads with head motions~\cite{sun2023diffposetalk, thambiraja2023_3diface, SadTalker_2023}, and expressive lip motion generation~\cite{chen2025diffusiontalker}. 
However, these diffusion-based frameworks still suffer from the limitation of fixed topology of the mesh; some of them, such as DiffPoseTalk~\cite{sun2023diffposetalk} require a morphable model, thus hindering the generalizability to other face datasets.

Despite these advancements, all current methods are constrained by their reliance on fixed mesh topologies. Our work addresses these limitations by proposing a novel framework that can animate and be trained with any face mesh, including real-world 3D scans, thereby enabling broader and flexible applications in 3D facial animation.

\subsection{3D Talking Heads Evaluation}
In 2D talking head generation, metrics like Peak Signal-to-Noise Ratio (PSNR), Structural Similarity Index Measure (SSIM), are commonly employed to evaluate visual fidelity. Meanwhile, lip synchronization can be measured using metrics such as Lip Sync Error Confidence (LSE-C), and Lip Sync Error Distance (LSE-D), which rely on visual features extracted by models like SyncNet~\cite{Chung16a_SyncNet, Wav2Lip}. Although these metrics effectively quantify the quality and synchronization of 2D videos, they are not directly applicable to 3D data, as the latter involve completely different measures and analyses. 
%For example, a frame might achieve high visual quality according to a 2D metric but still contain inconsistencies in its underlying 3D structure or lack temporal coherence across multiple viewpoints.
%Therefore, comprehensive evaluation metrics are required to assess the visual quality, structural accuracy, and temporal consistency of 3D-generated talking heads. 

Since the release of VOCA~\cite{VOCA2019}, the first high-quality open dataset for speech-driven 3D talking heads generation, effectively measuring the quality of the animations has been a challenge. Indeed, this task requires metrics capturing both the temporal aspect (mostly lip synchronization), and the non-Euclidean geometry of meshes in 3D space. To this aim, MeshTalk~\cite{richard2021meshtalk} proposed the Lip-Vertex Error (LVE) that measures the maximum quadratic error between the mouth vertices of the ground truth and predictions. However, LVE focuses on maximum error, which can be misleading: it may be influenced by a single inaccurate vertex, while the overall lip-sync quality remains high. Furthermore, the LVE metric suffers from ambiguity in defining mouth vertices, leading to inconsistent results across different studies due to varying lip vertex masks. This metric was then adopted in most subsequent works. CodeTalker~\cite{xing2023codetalker}, in addition to LVE, proposed the Face Dynamics Deviation (FDD) to measure upper-face movement quality. %In talking head synthesis though, the primary focus is on lip-syncing rather than upper-face movements. 
Despite natural and dynamic face movements are important for realism, the primary focus when generating 3D talking heads should be the accuracy and intelligibility of the speech. FaceDiffuser~\cite{FaceDiffuser_Stan_MIG2023} further introduced two additional metrics: Mean Vertex Error (MVE) and Diversity. MVE measures the maximum of the mean squared error between the ground truth and predictions, while Diversity assesses the variation in generated animations given the same audio and face. While these metrics provide insights into certain aspects of the animation, they still fall short in capturing the true quality of dynamic lip-syncing. To account for the temporal dynamics, Imitator~\cite{Thambiraja_2023_ICCV_imitator} introduced Dynamic Time Warping (DTW) to evaluate the quality of generated faces across all vertices and mouth vertices, yet the ambiguity of the masking factor for the mouth remains. 
On a similar trend, Nocentini \textit{et al}.~\cite{landmarks_3D_Nocentini_2023} proposed the Displacement Angle Discrepancy (DAE), which measures the angular difference between subsequent frames of ground-truth and generated animations. This helped in providing a coarse measure of the consistency of vertices motion. However, it does not tell much about the geometric similarity, as two similar motions might result from very different meshes. 
On a different perspective, SelfTalk~\cite{peng2023selftalk} proposed the Lip Readability Percentage (LRP) to assess the intelligibility of spoken words. However, it is an approximate measure based on the distance between mouth vertices. It is basically a thresholded version of the LVE. 
FaceTalk~\cite{aneja2023facetalk} showcased how other metrics on the rendering of talking heads can be efficiently used such as Lip Sync Error Distance (LSE-D)~\cite{Wav2Lip}, FID/KID~\cite{FID_KID_2023}, Face Image Quality Assessment~\cite{SDD-FIQA2021} and Video Quality Assessment~\cite{VQA_wu2023dover}. While providing interesting insights regarding the qualitative performance of the models, they may loose non-linear 3D contexts after projection in flat spaces and are only applcable for textured meshes.

Many of these metrics fail to effectively measure the quality of lip-syncing if considered alone. To address this, we thoroughly analyze their values and shortcomings, and propose a set of complementary metrics specifically designed to provide a comprehensive view of the quality of lip-syncing and overall generation, thus filling in a crucial gap in the current evaluation methodologies.

\section{Motion Synthesis with Space-time Features}\label{sec:approach}
\begin{figure}
    \centering
    \includegraphics[width=0.95\linewidth]{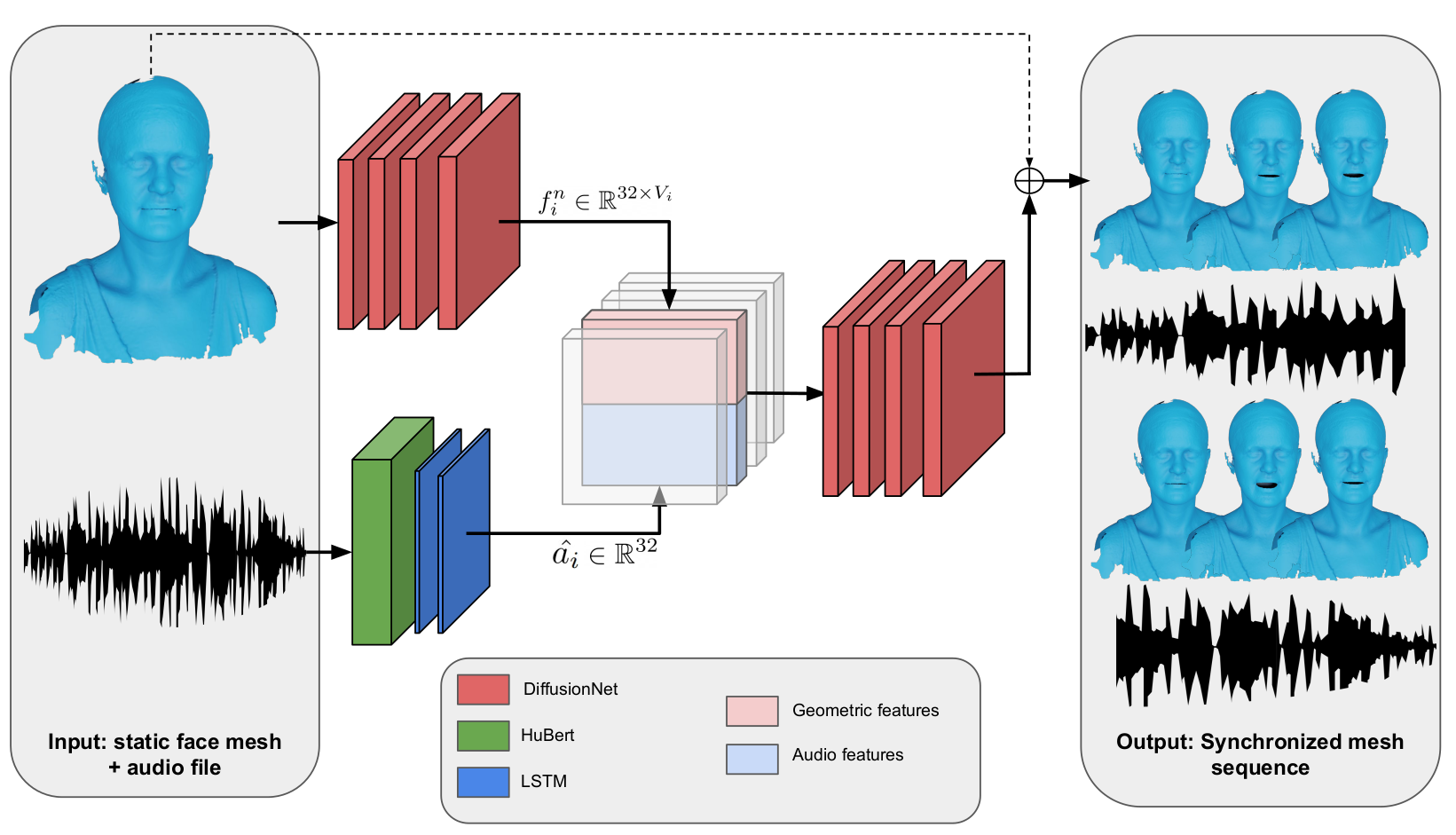}
    \caption{\textbf{Overview of the proposed deep architecture.} From a static mesh and an audio file, it computes time-dependent per-vertex features as a concatenation of geometric features $f_i^n$ and audio features $\hat{a_i}$. This learned signal over the mesh is used to learn a time-dependent displacement field, which produces the motion. At training time, the generated sequence is compared to the ground truth with different loss functions for the registered and unregistered cases. }%\textcolor{red}{The figure misses the pre-computed operators. Maybe a standalone figure?}}
    \label{fig:model_overview}
\end{figure}

An overview of the proposed method is illustrated in Figure~\ref{fig:model_overview}. In this section, we detail the different elements that compose our model and how they are combined. 
ScanTalk utilizes an Encoder-Decoder framework, which takes a neutral face mesh and an audio snippet as inputs, and outputs a sequence of per-vertex deformation fields. This time-dependent deformation fields, define how the neutral face mesh should be deformed to produce the animated sequence. The encoder consists of two modules: an audio encoder, which integrates a pretrained encoder with a bi-directional LSTM to extract audio features from the speech, and a DiffusionNet encoder that predicts surface descriptors from the neutral 3D face. These descriptors are replicated and concatenated with the audio features, and fed into a DiffusionNet decoder to produce the deformations to be applied to the neutral face. 

The above architecture is used both in the registered setting~\cite{nocentini2024scantalk3dtalkingheads} and the newly introduced unregistered one, but the training process, losses and evaluation metrics need to be adapted to each case. In Section~\ref{sec:protocols}, we describe the two settings along with the notations, in Section~\ref{sec:strategy} we detail the two strategies we used for training ScanTalk, and in Section~\ref{sec:proposedmetrics} we illustrate the evaluation metrics.
The specifics on the encoding and decoding operations are detailed, respectively, in Section~\ref{sec:encoder}, and Section~\ref{sec:decoding}. 

\subsection{Protocols and Notation}\label{sec:protocols}
ScanTalk was designed with the goal of animating meshes in arbitrary topologies. In our first proposal~\cite{nocentini2024scantalk3dtalkingheads}, any mesh, seen or unseen during training, could be animated at inference time. Yet, during training, meshes within each training sequence had a fixed and consistent mesh structure (registered setting): the sequences are said to be homogeneous. This allowed us to rely on losses that fully exploit the known vertex-wise correspondence. In the new, unregistered setting, this constraint is dropped. %Below we formalize the two protocols.

%\subsection{Notation}\label{sec:notation}
\paragraph{Registered Setting}
Let $L = \{(M_i^{gt}, m_i^{n}, A_i)\}_{i=0}^{N-1}$ represent the training set comprising $N$ samples, where $A_i$ is an audio file containing a spoken sentence, $M_i^{gt} = (m_i^0, \dots, m_i^{T_i-1}) \in \mathbb{R}^{T_i \times V_i \times 3}$ denotes a sequence of 3D faces (\textit{sharing the same mesh topology}) of length $T_i$, synchronized with the spoken sentence in $A_i$, and $m_i^{n} \in \mathbb{R}^{V_i \times 3}$ is a 3D neutral face. Here, $V_i$ is the number of vertices in the $i$-th 3D face sequence, and must remain consistent across each mesh in that sequence, along with vertex connectivity, defining the surface mesh topology. Note that in this setting, two different sequences $M_i^{gt}, M_j^{gt}$ might have different mesh topologies, \textit{i.e.}, $V_i \ne V_j$. This allowed us to train ScanTalk on multiple dataset despite different mesh structures. Our objective is then to learn a function that maps an audio input $A_i$ and a neutral 3D face $m_i^{n}$ to the ground-truth sequence $M_i^{gt}$:
\begin{equation}
\widehat{M}_i = ScanTalk(A_i, m_i^{n}) \approx M_i^{gt}.
\end{equation}
The topology of the meshes in the generated sequence $\widehat{M}_i$ is frame-wise consistent, and matches that of the input neutral face $m_i^{n}$. This holds both at training and inference. Since $m_i^{n}$ and the meshes in $M_i^{gt}$ have the same structure by construction, we can define losses that exploit the knowledge of the point correspondence (Section~\ref{subsec:strategy-reg}).

\paragraph{Unregistered Setting}
% \begin{wrapfigure}{r}{0.4\linewidth}
%     \vspace{-1cm}
%    \centering
%    \includegraphics[width=\linewidth]{img/reg_unreg_sequences.png}
%    \caption{\textcolor{blue}{Original vs. Remeshed VOCAset frames.}}
%    \label{fig:reg_unreg_sequences}
%    \vspace{-1.4cm}
% \end{wrapfigure}
In this setting, meshes in each sequence of the training set $L$ can be of arbitrary topology, that is, given a generic sequence $M_i^{gt} = (m_i^0, \dots, m_i^{T_i-1})$ each mesh $m_i^j, \; j=0, \cdots, T_i-1$ can have its own topology, \textit{i.e.}, $m_i^j \in \mathbb{R}^{V_i^j \times 3}$, $m_i^k \in \mathbb{R}^{V_i^k \times 3}$ with $V_i^j \ne V_i^k$ in general. Thus, we need different loss formulations that do not rely on vertex-wise correspondence within the training sequence (Section~\ref{subsec:strategy-unreg}). Similar to the registered case, both during training and inference, the topology of the meshes in the generated sequence $\widehat{M}_i$ is fixed, matching that of the neutral face $m_i^{n}$. However, this time can differ from that of the ground-truth sequence $M_i^{gt}$. 
%\textcolor{blue}{The difference between sample frames of a registered and an unregistered sequence can be observed in Figure~\ref{fig:reg_unreg_sequences}.}

\subsection{Encoding geometry and audio into a unified space}\label{sec:encoder}

ScanTalk requires an audio snippet $A_i$ and a 3D face in a neutral state $m_i^n$ as inputs. ScanTalk uses two distinct encoders for processing geometric and audio data, and later build a unified \textit{space-time} feature space encapsulating both information. The two modules and the resulting combination are detailed below.

\subsubsection{Face Mesh Encoder}
Several approaches are available for encoding face meshes, but traditional graph convolution-based models, like~\cite{gong2019spiralnet++, spiralconv_Lim_2019}, encounter limitations with varying graph structures. Thus, we employ DiffusionNet~\cite{sharp2021diffusion}, a discretization-agnostic encoder that has proven effective for encoding face meshes as in~\cite{Qin_2023_NFR}. DiffusionNet integrates multi-layer perceptrons (MLPs), learned diffusion, and spatial gradient features, offering a straightforward yet robust architecture for surface learning tasks. %It bypasses the need for complex operations like explicit surface convolutions.

In order to compute geometric features, DiffusionNet requires several precomputed operators, which are the \textit{Cotangent Laplacian} ($Cl$), \textit{Eigenbasis} ($Eb$), \textit{Mass Matrix} ($Mm$), and \textit{Spatial Gradient Matrix} ($SgM$). These features capture geometric properties of the face and facilitate information propagation across the surface domain. The Cotangent Laplacian, a discrete version of the Laplace-Beltrami operator, quantifies surface curvature and smoothness during the diffusion process. The Eigenbasis comprises eigenvectors derived from the Laplacian matrix, representing fundamental modes of variation on the surface. The Mass Matrix characterizes the distribution of mass or area across vertices, while the Spatial Gradient Matrix captures spatial derivatives of scalar functions defined on the surface. These operators enhance the architecture's robustness and flexibility, accommodating 3D faces of any topology. 
%making DiffusionNet suitable for diverse surface learning tasks. This architecture accommodates 3D faces of any topology, allowing for variations in the number and order of points.

Let $P_i^n = [Cl, Eb, Mm, SgM] = OP(m_i^{n})$ represent the precomputed features obtained by applying the above closed-form surface operators $OP$ to the 3D neutral face $m_i^{n}$. Note that these can be computed offline and do not need to be extracted multiple times as the neutral face to be animated remains fixed. The DiffusionNet Encoder $DN_e$, with latent size $h$, then processes a neutral 3D face mesh $m_i^{n}$ and the precomputed per-vertex features $P_i^n$ to extract \textbf{per-vertex descriptors} $f_i^{n}$ capturing intricate details of each vertex within the neutral 3D face:
\begin{equation}
f_i^{n} = DN_e(m_i^{n}, P_i^n) \in \mathbb{R}^{V_i \times h} .
\label{eq:per-vertex-descriptors}
\end{equation}

\subsubsection{Audio Encoder}
Following~\cite{facexhubert, FaceDiffuser_Stan_MIG2023}, the speech is encoded using a pre-trained \textit{HuBERT}~\cite{Hubert_audio_encoding} model
%which is a self-supervised speech representation learner utilizing an offline clustering step to provide aligned target labels for a BERT-like prediction loss. 
followed by a Linear layer, with which we obtain a per-frame audio representation:
\begin{equation}
a_i = SpeechEncoder(A_i) \in \mathbb{R}^{T_i \times (h/2)} .
\end{equation}

\noindent
To ensure temporal coherence in the speech representation, following the methodology outlined in~\cite{landmarks_3D_Nocentini_2023}, we concatenate the \textit{SpeechEncoder} with a Multilayer Bidirectional-LSTM, projecting the speech signal into a temporal latent representation:
\begin{equation}
\hat{a_i} = BiLSTM(a_i) \in \mathbb{R}^{T_i \times h} .
\end{equation}
The reason for choosing an LSTM over more complex models, \textit{e.g.}, Transformers, is that, in a speech, long range temporal dependencies do not impact much as the mouth shape changes are only influenced by the previous phonemes.

% \begin{figure}
%     \centering
%     \includegraphics[width=0.6\linewidth]{img/latent_new_notation.png}
%     \caption{The space-time feature field: the model learns a feature field over the static mesh that encapsulates in its structure spatial and temporal information. Red bars indicate the geometric features $(f_i^{n})_k$, while blue bars the audio features $\hat{a_i}$. The color gradient of the bars depicts how the geometric features (red) vary across vertices $k$, but remain fixed across time. The opposite occurs for audio features (blue), which are replicated for each vertex $k$, but vary at each frame $j$. For a sequence of length $T$, and a neutral mesh with $V$ vertices, the shape of the feature is $T \times 64 \times V$.}
%     \label{fig:latent_structure}
% \end{figure}

\subsubsection{Space-time feature field}\label{sec:space-time-field}
Properly combining geometric and audio features is of utmost importance, as it will form the base feature space for decoding the final talking head sequence. The main challenge is how to combine these features such that: \textit{(i)} the process can be done regardless of the mesh topology, and \textit{(ii)} audio features guide the motion synthesis irrespective of the underlying shape. 
%\begin{wrapfigure}{r}{0.50\linewidth}
%    \centering
%    \includegraphics[width=0.9\linewidth]{img/latent_new_notation.png}
%   \caption{Space-time feature field learned over the static mesh. Red bars: geometry features $(f_i^{n})_k$ (spatially varying, temporally fixed). Blue bars: audio features $\hat{a_i}$ (spatially replicated, temporally varying).}% Feature shape: $T \times 64 \times V$.}
%\label{fig:latent_structure}
%\end{wrapfigure}
To do so, we chose to concatenate per-vertex descriptors $f_i^{n}$ extracted from the neutral face with temporal latent vectors $\hat{a_i}^j$ from the Bidirectional-LSTM. Specifically, for each frame (or timestep) $j$, the same audio features $\hat{a_i}^j$ are replicated and concatenated for each vertex $k$. Conversely, the same geometric features of the neutral face $f_i^{n}$ are used for each frame $j$. More formally: 
%
% \begin{equation}
% (F_i^j)_k = (f_i^{n})_k \oplus \hat{a_i}^j \in \mathbb{R}^{h*2} , 
% \end{equation}
% \begin{equation}
% \forall \; k = 0, \dots, V_{i-1}; \;\;\; \forall \; j = 0, \dots, T_{i-1}.
% \end{equation}
%
\begin{align}
    (F_i^j)_k = (f_i^{n})_k \oplus \hat{a_i}^j \in \mathbb{R}^{h*2} 
    \qquad
    \forall \; k = 0, \dots, V_{i-1}; \;\;\; \forall \; j = 0, \dots, T_{i-1} , \label{eq:disp}
\end{align}

\noindent
where $k$ is a mesh vertex, $j$ is a frame in the sequence, and $i$ a sample in the dataset. This structure can be applied to neutral faces with arbitrary topology. Furthermore, since the geometric features are the same for each frame, the motion is guided by the audio features, 
We thus obtain a combined latent $F_i^j \in \mathbb{R}^{V_i \times h*2}$ that embeds both audio-related and geometry-related features. This allows us to propagate the dimensionality of the mesh, $V_i$, to the decoder, enabling the desired animation regardless of the mesh structure. A simple concatenation proved effective enough as compared to more elaborated strategies, \textit{e.g.}, attention mechanism. A quantitative analysis of different combination options are reported in the supplementary material.

\subsection{Decoding the talking sequence}\label{sec:decoding}
To predict the deformation sequence, we employ a DiffusionNet Decoder (essentially a reversed Encoder), denoted as $DN_d$. The decoder module receives $F_i^j$ and precomputed features $P_i^n$ derived from $m_i^{n}$, predicting the deformation of $m_i^{n}$ as:
\begin{equation}
\widehat{m_i}^j = DN_d(F_i^j, P_i^n) + m_i^{n} \in \mathbb{R}^{V_i \times 3} .
\end{equation}

\noindent
Here, $\widehat{m}_i^j$ represents the $j$-th frame of the predicted sequence. The entire generated sequence is defined by $\widehat{M}_i \in \mathbb{R}^{T_i \times V_i \times 3}$. To clarify the process, suppose we want to animate a 3D face with $V_i$ vertices, $V_i$ is arbitrary. 
The DiffusionNet encoder $DN_e$ processes each vertex of $m_i^n$ along with precomputed features $P_i^n$, producing a descriptor array of shape $[V_i, h]$. These descriptors are then concatenated with the descriptors $v_i$ from the audio processing pipeline. Each audio feature has the same size $h$ of the geometric features, and is \textit{replicated for each vertex}, resulting in a combined feature set of shape $[V_i, h*2]$ per audio frame. This combined feature set, along with the precomputed operators $P_i^n$ from the input mesh, is fed to the decoder. The decoder outputs a per-vertex displacement field from a time-dependent per-vertex descriptor field, yielding an output array of shape $[V_i, 3]$ per audio frame. Predicting a deformation field rather than reconstructing the actual geometry of the animated face offers advantages in terms of training efficiency and resulting animation as it simplifies the learning process by focusing solely on speech-related motion.

%ScanTalk predicts the \textit{deformation} of the neutral face $m_i^n$ rather than the actual face. This decision 
%aligns with previous works~\cite{Fan_Lin_Saito_Wang_Komura_faceformer_2022, facexhubert, peng2023selftalk, FaceDiffuser_Stan_MIG2023, Thambiraja_2023_ICCV_imitator}, 
%offers advantages in terms of training efficiency and resulting animation. Predicting face deformation simplifies the learning process by focusing solely on speech-related motion, rather than the entire face reconstruction. The decoder thus learns to predict .

\subsection{Training strategy}\label{sec:strategy}
The architecture of ScanTalk allows both registered and fully unregistered training. 
%This is different with respect to our previous version in~\cite{nocentini2024scantalk3dtalkingheads}, where meshes within a training sequence needed to adhere to a consistent topology (Section~\ref{sec:protocols}). %\rev{In the fully unregistered setting, this constraint is dropped, and each mesh $m_i^j$ of any sequence $i$ can be of different topology, \textit{even within the same sequence}.} 
However, due to its design, the generated deformation field maintains the same topology as the neutral face given as input for animation (which can be arbitrary). Expanding on the discussion in Section~\ref{sec:protocols}, this distinct feature ensures that even in the unregistered setting the model outputs a sequence of meshes that preserve the topology of the input neutral mesh, while being trainable with per-frame unregistered sequences. Thus, if a sequence is composed of $N$ meshes, each with a different topology, the generated sequence will maintain that of the neutral face. Implicitly, it performs a sort of dense registration, which is a valuable byproduct. This is achieved thanks to the Decoder, which requires knowledge of the precomputed features $P_i^n$ and number of vertices of the input mesh, ultimately defining its structure. %In the following, we report the training details for the two settings, \textit{i.e.}, registered and unregistered.

\subsubsection{Registered Setting}\label{subsec:strategy-reg}
In the registered setting, the ground-truth meshes \textit{within each sequence} must share a common topology. Yet, different sequences can have meshes of different structures. Nonetheless, it proved to be non-restrictive in practice as at inference time, any mesh can be animated even if its topology was unseen during training. Furthermore, the advantage is that one can exploit the additional knowledge of the mesh structure to impose specific losses or sophisticated constraints. We employed four different losses: 

%In a registered setting, the ground truth sequence is registered. As a result, the predicted and ground-truth sequences share the same topology, allowing for the definition of specific losses during training. 

\smallskip
\noindent
\textbf{Mean Squared Error Loss ($\mathcal{L}_{MSE}$)} minimizes the average vertex-wise Mean Squared Error (MSE) over a sequence of length $T_i$ between the ground truth mesh $M_i^{gt}$ and the model prediction $\widehat{M}_i$. The loss function is defined as:
\begin{equation}
    \mathcal{L}_{MSE} = \frac{1}{T_i-1}\sum_{j=0}^{T_i-1}\frac{1}{V_i-1}\sum_{k=0}^{V_i-1}\left \|(m_i^j)_k - (\widehat{m}_i^j)_k\right \|_2^2 .
\end{equation}

\noindent
\textbf{Masked Mean Squared Error Loss ($\mathcal{L}_{M}$)} is similar to $L_{MSE}$ but applied only to the mouth vertices. Here, $M_k$ is 1 if the vertex $k$ is a mouth vertex, and 0 otherwise.
\begin{equation}
    \mathcal{L}_{M} = \frac{1}{T_i-1}\sum_{j=0}^{T_i-1}\frac{1}{V_i-1}\sum_{k=0}^{V_i-1} M_k \left \|(m_i^j)_k - (\widehat{m}_i^j)_k\right \|_2^2 .
\end{equation}

\noindent
\textbf{Velocity Loss ($\mathcal{L}_{Vel}$)} minimizes the MSE between the motion of consecutive frames. It is intended to guide the generation by forcing similar per-frame dynamics. By defining the motion of consecutive frames for each vertex as:
%
% \begin{equation}
%     (d_i^j)_k = (m_i^j)_k - (m_i^{j-1})_k ,
%     \label{eq:perf-displ1} 
% \end{equation}
% \begin{equation}
%     (\widehat{d_i}^j)_k = (\widehat{m_i}^j)_k - (\widehat{m_i}^{j-1})_k ,
%     \label{eq:perf-displ2}
% \end{equation}
%
\begin{align}
    (d_i^j)_k = (m_i^j)_k - (m_i^{j-1})_k,
    \qquad
    (\widehat{d_i}^j)_k = (\widehat{m}_i^j)_k - (\widehat{m}_i^{j-1})_k,
     \label{eq:disp}
\end{align}

\noindent
the loss function can then be defined as:
\begin{equation}
    \mathcal{L}_{Vel} = \frac{1}{T_i-1}\sum_{j=1}^{T_i-1}\frac{1}{V_i-1}\sum_{k=0}^{V_i-1} \\ \left \|(d_i^j)_k - (\widehat{d_i}^j)_k\right \|_2^2 .
\end{equation}

\noindent
\textbf{Cosine Distance Loss ($\mathcal{L}_{Cos}$)} is similar to $\mathcal{L}_{Vel}$, but minimizes an angular measure instead of the MSE between per-frame motion vectors. It measures the cosine distance between the predicted and ground-truth displacements in~\eqref{eq:disp} relative to the previous frame. By defining the cosine distance as:
\begin{equation}
    C_d((d_i^j)_k, (\widehat{d_i}^j)_k) = 1 - \frac{(d_i^j)_k \cdot (\widehat{d_i}^j)_k}{\| (d_i^j)_k \|_2^2 \| (\widehat{d_i}^j)_k \|_2^2} ,
\end{equation}

\noindent
the loss function then becomes:
\begin{equation}
    \mathcal{L}_{Cos} = \frac{1}{T_i-1}\sum_{j=0}^{T_i-1}\frac{1}{V_i-1}\sum_{k=0}^{V_i-1} C_d((d_i^j)_k, (\widehat{d_i}^j)_k) . 
\end{equation}

\noindent
The above losses can be used under the assumption that each mesh within a sequence has a known, consistent topology. In the fully unregistered setting, this constraint is dropped. Thus, we need to define a different set of loss functions to train ScanTalk.

\subsubsection{Unregistered Setting}\label{subsec:strategy-unreg}
All methods in the literature for 3D talking head generation assume the training meshes have a known graph structure. In a fully unregistered training setting instead, each mesh in the training dataset might have its own topology, even within a given sequence. This unexplored scenario calls for the definition of alternative loss functions. In this work, we employ the \textbf{Chamfer distance} as loss, which is a well-established metric for estimating the difference between two generic point-clouds. It is defined as:
\begin{equation}
    d_{CD}(m_i^j, \widehat{m}_i^j) = \frac{1}{\widehat{V}_i}\sum_{l}\min_{k}\|(m_i^j)_k - (\widehat{m}_i^j)_l\|_2^2 + 
    \frac{1}{V_i}\sum_{k}\min_{l}\|(m_i^j)_k - (\widehat{m}_i^j)_l\|_2^2 ,
\end{equation}

\noindent
where the first term measures the average Euclidean distance between each vertex $l$ in the generated mesh $\widehat{m}_i^j$ and its corresponding nearest-neighbor $k$ in the ground-truth mesh $m_i^j$. The second term, instead, measures the average Euclidean distance between each vertex $k$ in the ground-truth mesh $m_i^j$ and its corresponding nearest-neighbor $l$ in the generated mesh $\widehat{m}_i^j$.
%for instance, $(m_i^j)_k$ is the $k$-th vertex of frame $j$ from ground truth sequence number $i$ and $(\widehat{m}_i^j)_k$ is its predicted counterpart. 
Then, we can average these distances along a sequence of length $T$ to obtain a dynamic Chamfer distance $\mathcal{L}^{CD}$ as:
\begin{equation}
    \mathcal{L}^{CD} =\frac{1}{T} \sum_{j=0}^T d_{CD}(m_i^j, \widehat{m}_i^j).
\end{equation}

Whereas other more sophisticated options exist to compute the discrepancy between generic meshes, in our experiments this loss proved efficient enough in comparison with other losses such as those based on varifolds or optimal transport~\cite{besnier_varifold_loss_2023, Croquet_Diff_Reg_OT_2021}. The latter are expensive to compute, even for static objects. For speech-driven dynamics, this is even more exacerbated, making them unsuitable.

\subsection{Proposed Evaluation Metrics}\label{sec:proposedmetrics}
In the realm of speech-driven 3D talking head animation, quantitatively assessing the quality of animated faces poses significant challenges. To define the proposed set of metrics, we mainly consider the registered case. %where ground-truth and generated meshes share the same topology regardless of the training scheme, 
The reason is twofold: first, 
%we are interested in assessing the quality of the animations. 
unregistered meshes require first to establish point correspondence, which is an error-prone process. Fully registered meshes allow for a more accurate error evaluation being the point correspondence known. Second, recent research showed how measuring the geometric error for unregistered 3D faces can be biased and uninformative~\cite{sariyanidi2023meta}. Nonetheless, to evaluate the performance of ScanTalk when used in the unregistered setting, we also present results using metrics that do not assume a known topology.

\subsubsection{Registered setting}\label{sect:registered-setting}
The most widely used metric for evaluating the accuracy of 3D Talking Head methods is the Lip-Vertex Error (LVE), which calculates the maximum quadratic error between the mouth vertices of the ground truth and the prediction. However, focusing on maximum error fails to capture the comprehensive quality of lip synchronization, as it may be skewed by a single inaccurate vertex, while overall lip-sync quality remains high. As a result, LVE is insufficient for a thorough evaluation of the generated faces and often produces rankings that do not align with qualitative assessments from visual inspections. To address this, we start with the assumption that evaluating lip-sync quality requires assessing the accuracy of lip \textit{movements}. In the registered setting for training and evaluation, we select six lip vertices (three per lip) for each topology (VOCAset, BIWI, Multiface) and measure the discrepancy between the dynamics of these vertices in the ground truth and the prediction. Specifically, we consider the 3D motion trajectory of these six points throughout the sequences (both generated and real). These can be interpreted as 3D signals, with each signal element corresponding to the $x$, $y$ and $z$ coordinates of the lip vertex in the respective face mesh of the sequence. This approach yields a collection of 3D signals that effectively capture lip movements, allowing for the application of time-series evaluation metrics. To assess the quality of lip synchronization, we propose leveraging two metrics commonly used for measuring distances between polygonal curves: the Discrete Fréchet Distance and Dynamic Time Warping, similar to~\cite{Thambiraja_2023_ICCV_imitator}. Furthermore, we advocate for assessing both the magnitude and angular deviations across subsequent frames. We thus define the \textit{Mean Squared Displacement Error} ($\delta_{M}$), and \textit{Cosine Displacement Error} $(\delta_{Cd})$. Given two sequences of points \( P = \{p_1, p_2, \ldots, p_n\} \) and \( Q = \{q_1, q_2, \ldots, q_n\} \), where \( p_i \) and \( q_j \) are points in \( \mathbb{R}^3 \), the following metrics are defined:

\smallskip
\noindent
\textbf{Dynamic Time Warping} (DTW)~\cite{muller2007dynamic} $\times 10^{-2}$ $(mm)$ measures the similarity between two sequences by allowing for non-linear alignments and different progression rates:
%
% \[
% DTW(P, Q) = \sqrt{ \min_{W} \sum_{(i,j) \in W} \| p_i - q_j \|^2 } ,
% \]
\begin{equation}
    DTW(P, Q) = \sqrt{ \min_{W} \sum_{(i,j) \in W} \| p_i - q_j \|^2 } ,
\end{equation}

\noindent
where the minimization is over all valid warping paths \( W \). This distance reflects the minimum cumulative cost of aligning the sequences, considering distortions in time.

\smallskip
\noindent
\textbf{Discrete Fréchet Distance} (DFD)~\cite{eiter1994computing} $\times 10^{-3}$ $(mm)$ measures the similarity between two sequences of points by considering their order. It is defined as:
%
% \[
% Frechet(P, Q) = \min_{\sigma} \max_{i} \| p_i - q_{\sigma(i)} \|_2 ,
% \]
\begin{equation}
    Frechet(P, Q) = \min_{\sigma} \max_{i} \| p_i - q_{\sigma(i)} \|_2 ,
\end{equation}

\noindent
where \( \sigma \) ranges over all the monotonic bijections between the indices of \( P \) and \( Q \). A monotonic bijection \( \sigma \) is a sequence of index pairs \( (i, j) \) such that \( 1 \leq i \leq n \), \( 1 \leq j \leq m \), and \( i \) and \( j \) are non-decreasing.

\smallskip
\noindent
% quantifies the discrepancy between two sequences of 3D points by
\textbf{Mean Squared Displacement} $\mathbf{\delta_{M}}$ $\times 10^{-6}$ $(mm)$ measures the magnitude of errors between the displacement vectors of two consecutive frames. It is defined as:
%
% \[
% \text{$Lip_{MSE}(P, Q)$} = \frac{1}{n-1} \sum_{i=2}^{n} \left\| (p_i - p_{i-1}) - (q_i - q_{i-1}) \right\|^2 ,
% \]
\begin{equation}
    \text{$\delta_{M}(P, Q)$} = \frac{1}{n-1} \sum_{i=2}^{n} \left\| (p_i - p_{i-1}) - (q_i - q_{i-1}) \right\|^2 ,
\end{equation}

\noindent
where \( p_i - p_{i-1} \) and \( q_i - q_{i-1} \) are the displacements between consecutive frames.

\smallskip
\noindent
\textbf{Cosine Displacement} $\mathbf{\delta_{Cd}}$ $(rad)$ evaluates the angular similarity between two trajectories by measuring the cosine distance between the displacement vectors of two consecutive frames. It is defined as:
%
% \[
% \text{$Lip_{Cos}(P, Q)$} = \frac{1}{n-1} \sum_{i=2}^{n} \left(1 - \frac{(p_i - p_{i-1}) \cdot (q_i - q_{i-1})}{\| p_i - p_{i-1} \| \| q_i - q_{i-1} \|} \right) ,
% \]
\begin{equation}
    \text{$\delta_{Cd}(P, Q)$} = \frac{1}{n-1} \sum_{i=2}^{n} \left(1 - \frac{(p_i - p_{i-1}) \cdot (q_i - q_{i-1})}{\| p_i - p_{i-1} \| \| q_i - q_{i-1} \|} \right) ,
\end{equation}

\noindent
where \((p_i - p_{i-1}) \cdot (q_i - q_{i-1})\) is the dot product of the displacement vectors, and \(\| p_i - p_{i-1} \|\) and \(\| q_i - q_{i-1} \|\) are their magnitudes.

\subsubsection{Unregistered setting} \label{subsubsec:unregistered_setting_method}
In this setting, the generated and ground-truth meshes differ in topology. In fact, all the generated meshes adhere to the input neutral mesh topology, which can differ from the ground-truth. In order to evaluate this scenario, since we do not have any known point-wise correspondence, we use three different shape distances: the Chamfer metric previously introduced, the Hausdorff distance (HD) and a kernel metric $\mathcal{L}^K$ on varifold representations of meshes introduced in~\cite{Charon_varifold_2013, kaltenmark_varifolds}. For two meshes $X$ and $\hat{X}$, it is essentially computed by blending areas $(a_f)$, centers $(c_f)$ and normals $(\vec{n}_f)$ of triangles faces with kernels $k_p$ and $k_n$:
\begin{equation}
    \mathcal{L}^{K} = \langle \mu_{X}, \mu_{X} \rangle + \langle \mu_{\hat{X}}, \mu_{\hat{X}} \rangle
    - 2\langle \mu_{X}, \mu_{\hat{X}} \rangle ,
\end{equation}

\noindent
with: 
\begin{equation}
    \langle \mu_X, \mu_{\hat{X}} \rangle = \sum_{f\in F(X)}\sum_{\hat f\in F(\hat{X})} a_f a_{\hat{f}} k_p(c_f,c_{\hat{f}})k_n(\vec{n}_{f},\vec{n}_{\hat{f}}) .
\end{equation}

\noindent 
The unoriented varifold metric is characterized by a scale $\sigma$ for the position kernel $k_p$:
\begin{align}
    k_p : \begin{cases}
        \mathbb{R}^3 \times \mathbb{R}^3 &\rightarrow \mathbb{R} \\
        (x, \hat{x}) &\mapsto \exp \left(\frac{\|x - \hat{x}\|}{\sigma^2} \right)
    \end{cases} 
    \qquad
    k_n:\begin{cases}
        \mathbb{S}^2 \times \mathbb{S}^2 &\rightarrow \mathbb{R}\\
        (\vec{n}_x, \vec{n}_{\hat{x}}) &\mapsto \langle \vec{n}_x, \vec{n}_{\hat{x}} \rangle^2 .
    \end{cases}
\end{align}

\section{Experiments}\label{sec:results}
In the following sections, we first outline the experimental methodology adopted for both registered and unregistered settings (Section~\ref{sec:experiments}). We then present a comprehensive quantitative evaluation of our approach in Section~\ref{sec:registered} and Section~\ref{sec:unregistered}. We also present the results of a user study in Section~\ref{sec:user}. Our models are trained and evaluated on three publicly available datasets: VOCAset~\cite{VOCA2019}, BIWI~\cite{BIWI_2010}, and Multiface~\cite{wuu2022multiface}. Additional details on these datasets can be found in the supplementary material.

\subsection{Experimental Protocols}\label{sec:experiments}
\textbf{Registered setting}. In this setting, all meshes of a given sequence are aligned to a common topology. This allows using a supervised loss such as the $\mathcal{L}_{MSE}$, $\mathcal{L}_{M}$, $\mathcal{L}_{Vel}$ and $\mathcal{L}_{Cos}$.  Even though the topology may vary from one sequence to another, when training a single model on several registered datasets the only requirement is to rigidly align and scale the data with a reference. For our experiments, we rigidly aligned all the meshes with the VOCAset meshes.

% \begin{wrapfigure}{r}{0.35\linewidth}
%     \vspace{-0.7cm}
%    \centering
%    \includegraphics[width=\linewidth]{img/same_face_2discretizations_cropped.png}
%    \caption{VOCAset mesh (left) and remeshed version (right) for unregistered training.}
%    \label{fig:remeshed_id}
%    \vspace{-1.3cm}
% \end{wrapfigure}
\noindent
\textbf{Unregistered setting}. Here, we replace $\mathcal{L}_{MSE}$ with the dynamic Chamfer distance $\mathcal{L}_{CD}$. 
To showcase the results of a training procedure on unregistered meshes, we chose to randomly remesh VOCAset by upsampling and downsampling the meshes.% as depicted in Figure~\ref{fig:remeshed_id}. 
In doing so, each mesh of each sequence has its own structure, different from all the others. 
For other settings, such as training directly on scans, one should take care about rigid alignment with the training data and the existence of a clear aperture for the mouth. Also, downsampling the training meshes can make the training faster.

\subsection{Registered Training}\label{sec:registered}
Table~\ref{tab:new_metrics} shows the performance of ScanTalk when trained with various loss combinations across all three registered datasets in a multidataset setting. The table reports both the proposed metrics and the standard ones. We compare against state-of-the-art methods and vanilla ScanTalk trained separately on each dataset. At a first glance, it can be easily noted that none of the compared approaches outperforms the others with respect to all the metrics. In fact, while standard metrics (LVE, MVE, FDD) mostly focus on geometric measures, they do not explicitly account for the dynamics of the animation. The proposed ones instead (DTW, DFD, $\delta_{M}$, $\delta_{Cd}$) provide such information. This points towards the idea that there is still large room for improvements, but without proper measurements such assessment is difficult.    

\begin{table*}[h!]
\vspace{-0.5cm}
\centering
\caption{Results across different datasets. MV stands for mask+velocity, MVC for mask+velocity+cosine. State-of-the-art models are trained on a single dataset, Scantalk is trained both on single (sd) and multi-dataset (md) settings.}
\label{tab:new_metrics}
\resizebox{\textwidth}{!}{
        \begin{tabular}{@{}l@{}c@{\hspace{0.1cm}}c@{\hspace{0.1cm}}c@{\hspace{0.1cm}}c@{\hspace{0.1cm}}c@{\hspace{0.1cm}}c@{\hspace{0.1cm}}c@{\hspace{0.3cm}}|c@{\hspace{0.1cm}}c@{\hspace{0.1cm}}c@{\hspace{0.1cm}}c@{\hspace{0.1cm}}c@{\hspace{0.1cm}}c@{\hspace{0.1cm}}c@{\hspace{0.3cm}}|c@{\hspace{0.1cm}}c@{\hspace{0.1cm}}c@{\hspace{0.1cm}}c@{\hspace{0.1cm}}c@{\hspace{0.1cm}}c@{\hspace{0.1cm}}c@{}}
        & \multicolumn{7}{c}{\textbf{\Large VOCAset}} & \multicolumn{7}{c}{\textbf{\Large BIWI\textsubscript{6}}} & \multicolumn{7}{c}{\textbf{\Large Multiface}} \\
        \toprule
        & LVE & MVE & FDD & DTW & DFD & $\delta_{M}$ & $\delta_{Cd}$ & LVE & MVE & FDD & DTW & DFD & $\delta_{M}$ & $\delta_{Cd}$ & LVE & MVE & FDD & DTW & DFD & $\delta_{M}$ & $\delta_{Cd}$ \\
        \midrule
        VOCA & 6.99 & 0.98 & 2.66 & 1.77 & 7.41 & 1.39 & 1.02 & 5.74 & 2.59 & 41.5 & 1.60 & 7.89 & 1.47 & 0.67 & 4.92 & 2.76 & 55.78 & 1.56 & 6.51 & 0.86 & 1.23 \\
        FaceFormer & 6.12 & 0.93 & \underline{2.16} & 1.33 & 5.39 & 0.86 & 0.58 & 4.08 & 2.16 & 37.1 & 1.56 & 6.61 & \underline{0.87} & \textbf{0.55} & 2.45 & \textbf{1.45} & \textbf{20.24} & \textbf{0.87} & \underline{4.13} & \textbf{0.22} & \textbf{0.73} \\
        FaceDiffuser & 4.35 & 0.90 & 2.43 & 1.73 & 6.83 & 0.81 & 0.60 & \underline{4.02} & 2.12 & 39.6 & 1.55 & \underline{6.50} & \textbf{0.85} & \underline{0.56} & 3.55 & 2.39 & \underline{29.16} & 1.45 & 5.24 & 0.28 & 0.77 \\
        CodeTalker & \underline{3.55} & \underline{0.89} & 2.26 & 1.33 & 5.66 & 0.80 & \underline{0.55} & 5.19 & 2.64 & \textbf{20.6} & 1.50 & 6.61 & 0.91 & 0.59 & 4.09 & 2.38 & 47.90 & 1.44 & 5.95 & 0.55 & 0.96 \\
        SelfTalk & 5.61 & 0.91 & 2.32 & 1.25 & 5.43 & \underline{0.72} & 0.56 & \textbf{3.63} & \underline{2.06} & \underline{35.5} & 1.60 & 6.93 & 1.03 & 0.66 & \textbf{2.28} & 1.90 & 37.43 & 0.96 & 4.18 & 0.25 & 0.75 \\
        \midrule
        ScanTalk (md) & 6.38 & 0.99 & 2.10 & 1.39 & 5.71 & 0.75 & 0.60 & 4.04 & \textbf{2.05} & 40.0 & \underline{1.47} & \textbf{6.48} & 0.97 & 0.60 & \underline{2.43} & \underline{1.67} & 32.20 & \underline{0.90} & \textbf{4.08} & \underline{0.23} & 0.74 \\
        +MV (md)& 7.03 & 1.09 & \underline{1.47} & \textbf{1.14} & \textbf{5.16} & 0.78 & 0.58 & 4.75 & 2.22 & 39.4 & 1.51 & 6.74 & 1.09 & 0.63 & 2.54 & 1.89 & 15.97 & 0.99 & 4.21 & 0.27 & 0.77 \\
        +MVC (md) & 7.05 & 1.07 & \textbf{1.29} & \underline{1.19} & \underline{5.25} & \textbf{0.72} & \textbf{0.52} & 4.44 & 2.09 & 36.6 & \textbf{1.46} & 6.71 & 1.01 & 0.61 & 2.37 & 1.71 & 16.00 & 0.93 & 4.71 & 0.33 & \underline{0.74} \\
        %+CH (md) & 6.57 & 1.01 & 2.31 & 1.53 & 6.24 & 0.77 & 0.59 & 4.30 & 2.14 & 34.9 & 1.54 & 6.75 & 0.99 & 0.62 & 3.71 & 2.52 & 110.6 & 1.36 & 5.64 & 0.35 & 0.74 \\
        \midrule
        ScanTalk (sd) & \textbf{3.01} & \textbf{0.86} & 2.40 & 1.20 & \underline{5.25} & 0.73 & \underline{0.55}  & 4.65 & 2.14 & 36.0 & 1.56 & 7.22 & 1.05 & 0.63 & 2.65 & 1.87 & 64.43 & 0.96 & 4.27 & 0.57 & 0.93 \\
        \bottomrule
        \end{tabular}}
        \vspace{-0.2cm}
\end{table*}

\begin{figure}[!ht]
    \centering
    \includegraphics[width=0.95\textwidth]{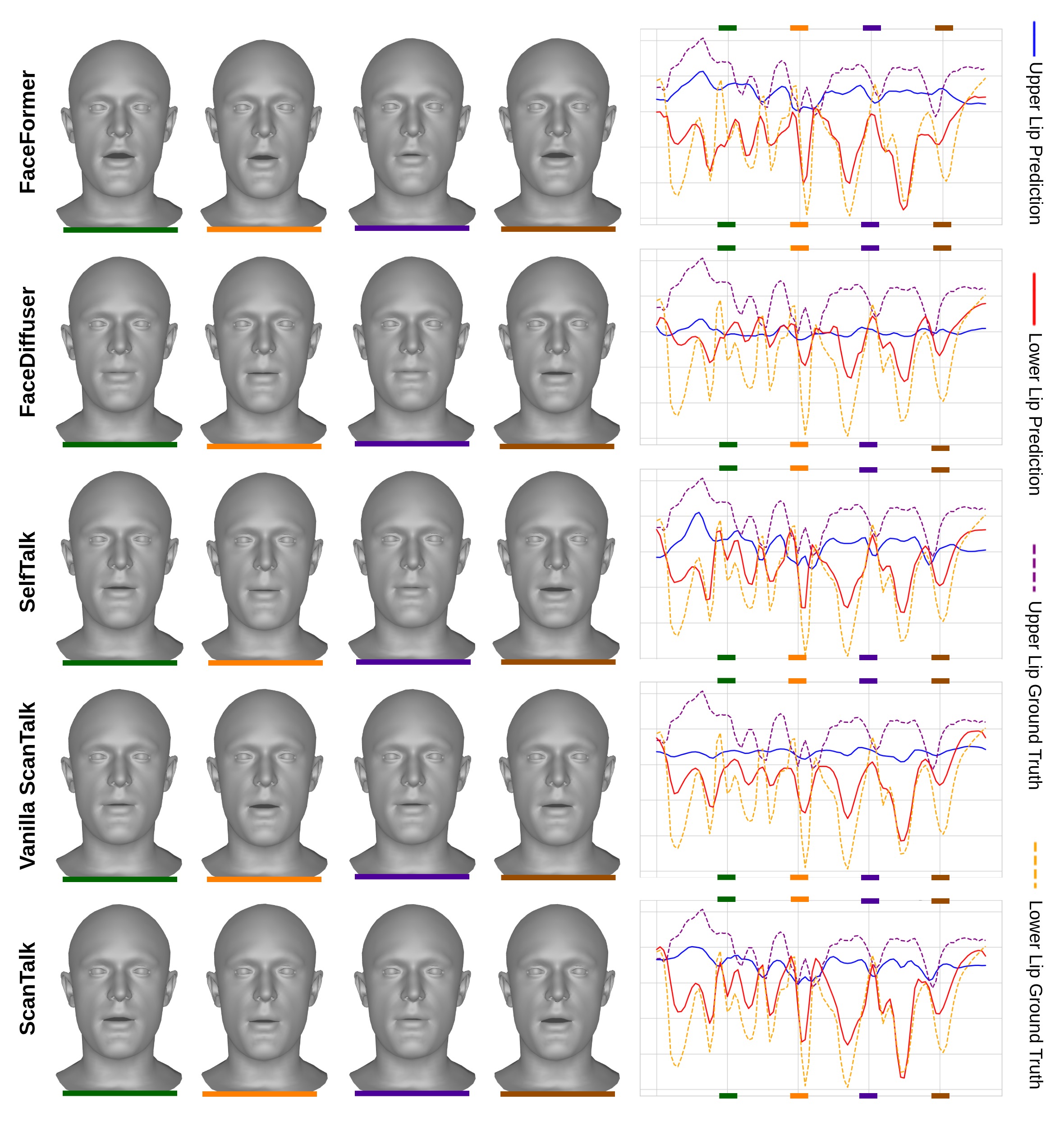}
    \caption{Comparison of ScanTalk variants and state-of-the-art methods on a VOCAset test sequence, focusing on vertical lip movements ($y$-coordinate). Each row shows visual outputs and corresponding lip-sync plots. The $y$-coordinate captures the main dynamics of lip motion. ScanTalk benefits from additional loss terms, yielding more accurate and expressive lip movement. Here same colors represent same timestep.}
    %\caption{Comparison on VOCAset test sequence. Each row shows rendered frames alongside lip motion plots ($y$-coordinate of upper and lower lip) over time.}
    %\vspace{-0.5cm}
    \label{fig:lip_charts}
\end{figure}

One interesting result that supports our claim is that in Table~\ref{tab:new_metrics} vanilla ScanTalk performs better than ScanTalk trained with the additional losses (masked, velocity and cosine) in terms of geometric measures such as LVE. However, if one looks at the charts in Figure~\ref{fig:lip_charts}, which reports both visual results and the generated $y$-coordinate movements of the upper and lower lips against ground truth, can clearly observe that the dynamics of the animation is better modeled when the additional losses are used. This is reflected in lower DTW, DFD, $\delta_{M}$ and $\delta_{Cd}$ measures in Table~\ref{tab:new_metrics}, which represent a quantitative alternative to the qualitative results in the charts.

\begin{figure*}[!ht]
    \centering
    \begin{subfigure}[t]{0.38\textwidth}
        \centering
        \includegraphics[width=\linewidth]{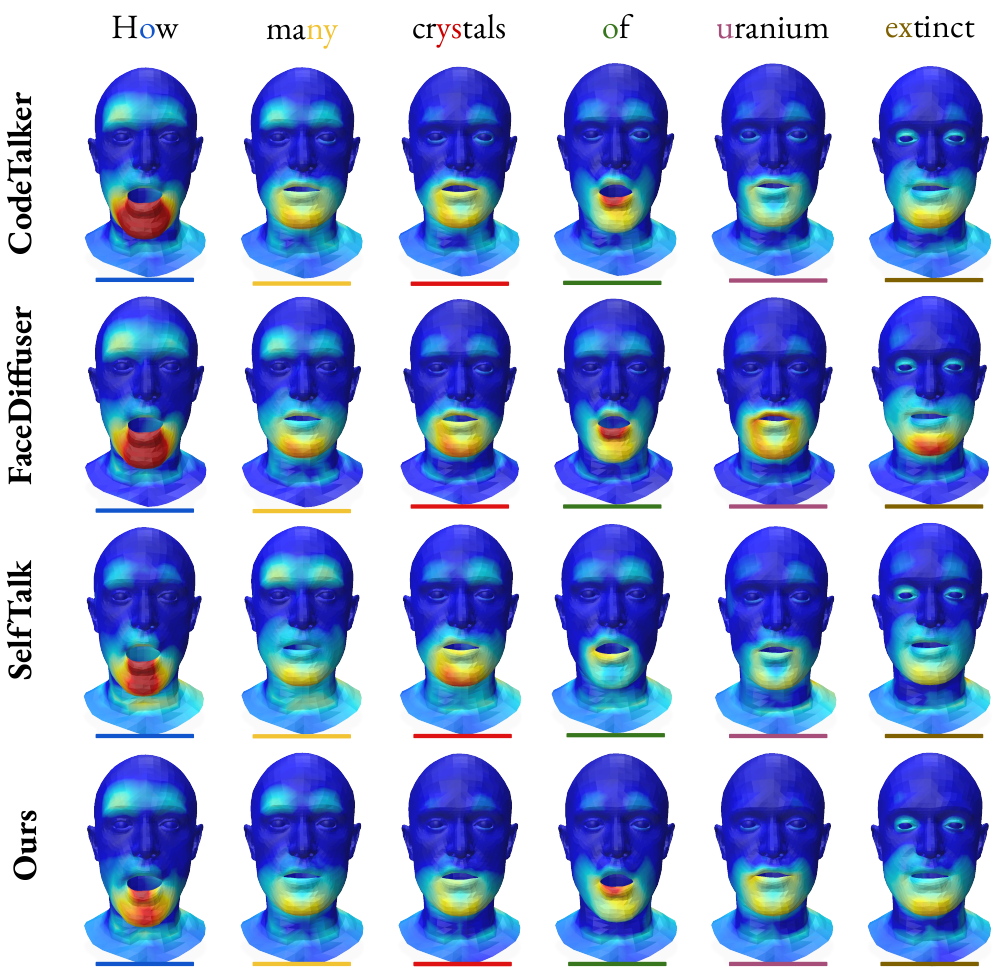}
        \caption{Comparison of ScanTalk with SOTA methods on VOCAset. Blue/red hues = low/high error.}
        \label{fig:heatmap_error}
    \end{subfigure}
    \hfill
    %\vline
    \begin{subfigure}[t]{0.6\textwidth}
        \centering
        \includegraphics[width=\linewidth]{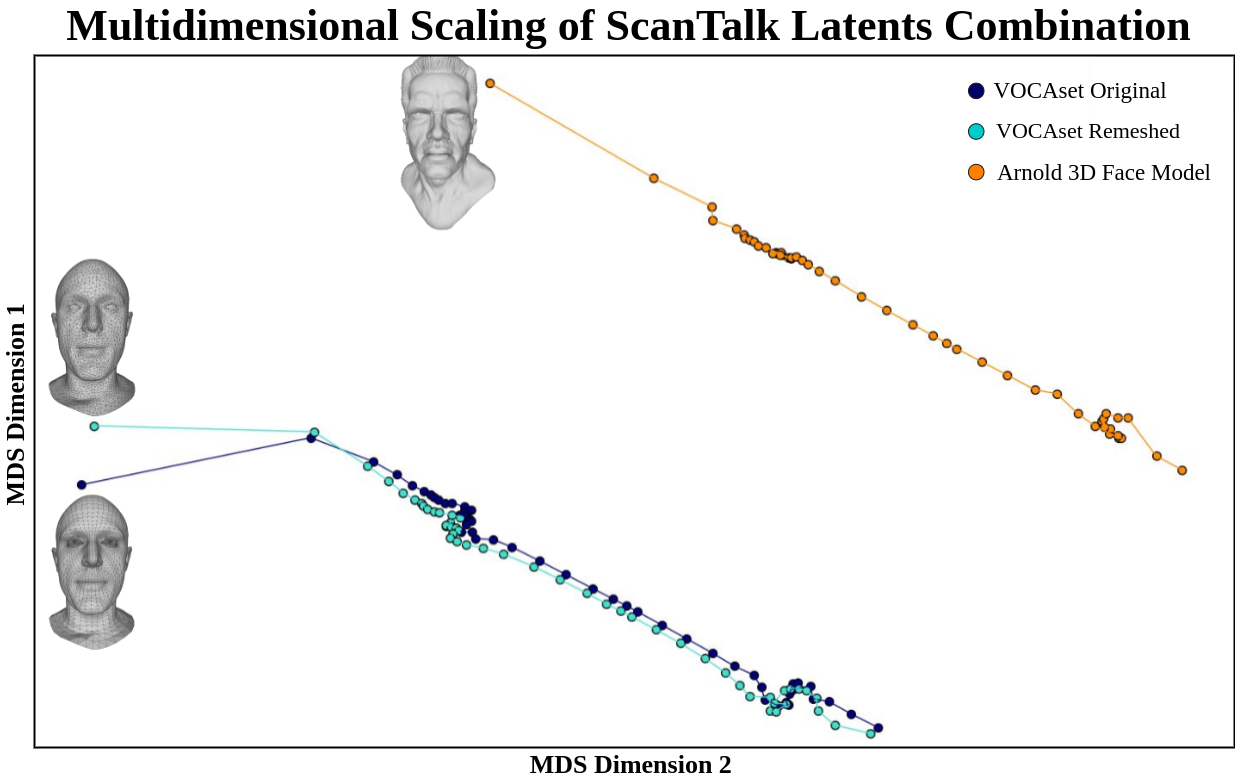}
        \caption{MDS visualization of latent combinations. Same identity-audio pairs show highly similar representations, despite different parameterization. Different identities also exhibit similar MDS trajectories.}
        \label{fig:mds}
    \end{subfigure}
    \caption{Visualization results comparing error maps and multidimentional scaling.}
    \label{fig:combined_figure}
    \vspace{-0.3cm}
\end{figure*}

Another notable observation from Table~\ref{tab:new_metrics} is that while standard metrics suggest that training vanilla ScanTalk on a single dataset yields better performance, our proposed metrics reveal the opposite: the multi-dataset training setting leads to improved performance, even on individual datasets. %This highlights the adaptability and effectiveness of our approach in capturing more generalizable lip motion patterns.
Overall, ScanTalk trained in a multidataset setting performs favorably with respect to state-of-the-art models, mostly on VOCAset, with the newly introduced metrics providing a more complete evaluation. For example, in VOCAset, FaceDiffuser and CodeTalker report the lowest LVE and MVE. However, their motion dynamics is quite compromised when looking at the generated animations. To support this, Figure~\ref{fig:lip_charts} reports their motion charts for a randomly picked sequence of VOCAset. The generated motion fails to a large extent in following that of the ground-truth. Indeed, this is again reflected in worse DTW, DFD and $\delta$ measures. ScanTalk trained with the additional losses, instead, better captures the motion dynamics. This improvement aligns with our expectations, as the new losses were specifically designed to enhance dynamic animation realism, a nuance that the standard metrics failed to capture. This comparison highlights the limitations of standard metrics, demonstrating their inadequate performance. In contrast, the new metrics we propose offer a significantly improved assessment of lip-sync quality. Heatmaps in Figure~\ref{fig:heatmap_error}, which show the MSE between the ground truth and predictions on VOCAset, qualitatively reveal that ScanTalk performs comparably with other state-of-the-art methods.

In $BIWI_6$ and Multiface performance vary more. Specifically, ScanTalk surpasses the state-of-the-art method on $BIWI_6$ in DTW and DFD, but not in $\delta_{M}$ and $\delta_{Cd}$. For Multiface, ScanTalk outperforms state-of-the-art methods for the DFD metric and performs comparably in terms of other metrics. These observations align with findings from~\cite{nocentini2024scantalk3dtalkingheads}, and we attribute this performance discrepancy to the significant differences among the three registered datasets. VOCAset features static faces with only mouth movements, while $BIWI_6$ and particularly Multiface involve notable head movements in their sequences. These dataset variations contribute to the model's reduced performance as they introduce a further level of complexity to be modeled in a unified training scheme. 

To evaluate the impact of such head rotations and the robustness of ScanTalk, we performed a specific test. We computed the relative difference $\Delta_{R}$ between a metric $M$ (\textit{e.g.}, LVE or others) computed on the aligned ($Original$) test set of VOCASET and when applying random rigid rotations by a certain angle ($Rotated$):
\begin{equation}
\Delta_{R} = \frac{M(Rotated) - M(Original)}{M(Original)} .
\end{equation}

The results in Table~\ref{tab:rotation_robustness} provide clear insights of the robustness of ScanTalk with respect to pose variation at test time. We observe that the error increase remains relatively modest even under significant rotations of the input mesh. When analyzing the LVE and MSE across rotations of up to $\pm 30^\circ$, the relative performance degradation is especially small for rotations around the $Y$ (yaw) and $Z$ (roll) axes. 
\begin{wraptable}[10]{r}{0.4\linewidth}
\vspace{-0.6cm}
\caption{Increase (\%) in LVE and MSE after applying random test-time rotations on VOCAset.}
\label{tab:rotation_robustness}
\resizebox{\linewidth}{!}{%
\begin{tabular}{lccc@{\hskip 10pt}ccc}
\toprule
\multicolumn{1}{c}{} & \multicolumn{3}{c}{LVE} & \multicolumn{3}{c}{MSE} \\ 
\cmidrule(lr){2-4} \cmidrule(lr){5-7}
Angle (deg.) & $X$ & $Y$ & $Z$ & $X$ & $Y$ & $Z$ \\ 
\midrule
$\pm 2^\circ$  & 0.4 & 0.1 & 0.1 & 0.1 & 0.1 & 0.1 \\
$\pm 5^\circ$  & 1.4 & 0.4 & 0.2 & 0.2 & 0.1 & 0.1 \\
$\pm 10^\circ$ & 4.2 & 1.0 & 0.5 & 0.4 & 0.3 & 0.8 \\
$\pm 20^\circ$ & 5.4 & 2.2 & 2.0 & 1.5 & 1.0 & 1.6 \\
$\pm 30^\circ$ &14.9 & 4.8 & 5.0 & 3.8 & 2.6 & 2.3 \\
\bottomrule
\end{tabular}}
\end{wraptable}
As we see in Table~\ref{tab:rotation_robustness}, the LVE error increases more significantly with larger rotations, especially along the $X$ (pitch) axis. The figure also shows the $XYZ$ coordinate system used to define rotation axes, overlaid on a sample VOCAset face, which helps interpret the directional impact of each rotation.
This stability indicates that the model is not overfitting to rigid, canonical head poses seen during training, but rather learns representations that are invariant to moderate pose shifts. In this sense, it acts as a form of data augmentation, which improves the overall robustness of the model. More details about this experiment can be found in the supplementary material.
\vspace{-0.5cm}
\vspace{-0.5cm}
Finally, to illustrate the consistency of the learned space-time feature space, in Figure~\ref{fig:mds} we present the Multidimensional Scaling (MDS)~\cite{torgerson1952multidimensional} projection of three sequences of latent combinations from ScanTalk, each generated with the same audio snippet but different faces to animate. The results show that when two faces share the same identity but differ in topology, their MDS sequences are very similar, nearly overlapping. In contrast, when animating different identities with the same audio, the MDS paths exhibit similar patterns but occupy distinct positions in space. These results underscore ScanTalk capability to animate faces independently of topology, focusing on the face identity. From Table~\ref{tab:new_metrics}, it is evident that ScanTalk, when trained with the Chamfer loss in a registered setting, yields results comparable to those obtained using registered losses in the same setting. This comparison underscores the robustness and adaptability of our framework.
\subsection{User study}\label{sec:user}
\begin{wraptable}[8]{r}{0.45\linewidth}
\vspace{-0.9cm}
\centering
\caption{User study on lip-sync quality: percentage of ratings per method across five categories.}
\resizebox{\linewidth}{!}{%
\begin{tabular}{lccccc}
\toprule
 & \textit{VP}(\%) & \textit{P}(\%) & \textit{F}(\%) & \textit{G}(\%) & \textit{VG}(\%)\\ 
\midrule
FaceFormer     & 14.9 & 30.6 & 25.5 & 24.5 & 4.5 \\
FaceDiffuser   & 40.1 & 39.9 & 14.3 & 5.7  & 0.0 \\
CodeTalker     & 1.2  & 16.3 & 31.4 & 30.2 & 20.9 \\
SelfTalk       & 0.0  & 4.5  & 30.9 & 45.5 & 19.1 \\
ScanTalk       & 1.3  & 10.3 & 26.0 & 38.5 & 23.9 \\
\bottomrule
\end{tabular}}
\label{tab:user}
%\vspace{-0.3cm}
\end{wraptable}

With the goal of validating our claim regarding the need for additional metrics to measure the realism of animations, we designed a user-based study so to relate the subjective perception to the quantitative metrics described in Section~\ref{sect:registered-setting} and reported in Table~\ref{tab:new_metrics}. In particular, 14 participants were shown 60 animations (all in FLAME topology, so to avoid potential biases) and asked them to evaluate the lip-sync quality and realism using a five-point scale: \textit{Very Poor (VP)}, \textit{Poor (P)}, \textit{Fair (F)}, \textit{Good (G)}, and \textit{Very Good (VG)}. We compared ScanTalk against 4 other state-of-the-art models~\cite{Fan_Lin_Saito_Wang_Komura_faceformer_2022, peng2023selftalk, FaceDiffuser_Stan_MIG2023, xing2023codetalker}. Note that we did not conduct the usual A/B study where users are asked which result they like more among multiple options; each animation is here given an independent score. The goal of this design is twofold: to assess to what extent \emph{(i)} subjective perception correlates with the quantitative metrics, and \emph{(ii)} the proposed dynamic losses affect the perceived realism with respect to other approaches.

Table~\ref{tab:user} summarizes the user study results, showing the distribution of scores assigned to each method. Overall, the findings reveal a strong alignment between subjective perception and our proposed metrics, with ScanTalk receiving the highest proportion of \textit{Very Good} ratings. Notably, FaceDiffuser, despite achieving a lower LVE than both ScanTalk and SelfTalk, received significantly lower user ratings, highlighting the inadequacy of standard  metrics in capturing perceived visual quality. In contrast, the user scores show a clear correlation with our proposed metrics, underlying their effectiveness in capturing the quality of lip motion dynamics.

\subsection{Unregistered Setting}\label{sec:unregistered} 
Training on raw scans directly is too challenging for several reasons. Teeth, hair, occlusions, and other physical attributes can introduce significant interference in the signal, making it difficult for the model to accurately learn lip movements. While holes and noise in the raw scans are generally manageable, features like teeth or hair can be problematic. The model typically collapses or overfits unless extra information is provided through multi-modal approaches for instance. However, our approach is still a significant step forward regarding the robustness and generalizability of deep learning-based 3D talking heads.
To support our claim that performance remains almost unchanged in an unregistered setting and to mimic raw scans attributes, we provide results of ScanTalk when trained on the three datasets where these undergo several types of alterations. In particular, we measure the performance when the training datasets are Registered (Rgt), Remeshed (Rmsh) by subdividing the mesh and then decimating it, randomly creating holes (H) or adding Gaussian Noise (GN) to vertex positions. More details on the data preparation are given in the supplementary material.
\begin{table*}[h!]
\vspace{-0.5cm}
\centering
\caption{Results across different datasets for the unregistered setting in different scenarios. All models are trained on the three datasets with different alteration: Registered (Rgt), Remeshed (Rmsh), randomly creating holes (H) and adding Gaussian Noise (GN) to vertex positions.}
\label{tab:newmetrics}
\resizebox{\textwidth}{!}{
        \begin{tabular}{@{}l@{}c@{\hspace{0.1cm}}c@{\hspace{0.1cm}}c@{\hspace{0.1cm}}c@{\hspace{0.1cm}}c@{\hspace{0.1cm}}c@{\hspace{0.1cm}}c@{\hspace{0.3cm}}|c@{\hspace{0.1cm}}c@{\hspace{0.1cm}}c@{\hspace{0.1cm}}c@{\hspace{0.1cm}}c@{\hspace{0.1cm}}c@{\hspace{0.1cm}}c@{\hspace{0.3cm}}|c@{\hspace{0.1cm}}c@{\hspace{0.1cm}}c@{\hspace{0.1cm}}c@{\hspace{0.1cm}}c@{\hspace{0.1cm}}c@{\hspace{0.1cm}}c@{}}
        & \multicolumn{7}{c}{\textbf{\Large VOCAset}} & \multicolumn{7}{c}{\textbf{\Large BIWI\textsubscript{6}}} & \multicolumn{7}{c}{\textbf{\Large Multiface}} \\
        \toprule
        Trained on& LVE & MVE & FDD & DTW & DFD & $\delta_{M}$ & $\delta_{Cd}$ & LVE & MVE & FDD & DTW & DFD & $\delta_{M}$ & $\delta_{Cd}$ & LVE & MVE & FDD & DTW & DFD & $\delta_{M}$ & $\delta_{Cd}$ \\
        \midrule
        Rgt (MSE) & 6.38 & 0.99 & 2.10 & 1.39 & 5.71 & 0.75 & 0.60 & 4.04 & 2.05 & 40.0 & 1.47 & 6.48 & 0.97 & 0.60 & 2.43 & 1.67 & 32.20 & 0.90 & 4.08 & 0.23 & 0.74 \\
        Rgt (CH) & 6.57 & 1.01 & 2.31 & 1.53 & 6.24 & 0.77 & 0.59 & 4.30 & 2.14 & 34.9 & 1.54 & 6.75 & 0.99 & 0.62 & 3.71 & 2.52 & 110.6 & 1.36 & 5.64 & 0.35 & 0.74 \\
        Rmsh (CH) & 7.13 & 1.43 & 8.22  & 1.49 & 6.04 & 0.81 & 0.71 & 5.59 & 2.37 & 28.2 & 1.90 & 8.20 & 1.24 & 0.69  &5.51 & 2.86 & 125.1& 1.76 & 6.41 &  0.33& 0.77 \\
        Rmsh+H (CH) & 7.13 & 1.37 & 9.94 & 1.73 & 6.80 & 0.93 & 0.71 & 5.45 & 2.21 & 37.1 & 1.75 & 7.72 & 1.04 & 0.66 & 5.69 & 3.00 & 26.1 & 1.66 & 6.43 & 0.41 & 0.79 \\
        Rmsh+H+GN (CH) & 7.91 & 1.80 & 39.0 &  1.75 & 6.57 & 0.88 & 0.68 & 4.86 & 2.23 & 37.2 & 1.63 & 7.19 & 1.04 & 0.66 & 5.75&  3.11& 125.0&  1.66& 6.44 & 0.42 & 0.78 \\
        \bottomrule
        \end{tabular}}
        \vspace{-0.2cm}
\end{table*}
 As shown in Table~\ref{tab:newmetrics}, when ScanTalk is trained on unregistered datasets, it still achieves strong performance on registered test data. Although some degradation in results occurs as data quality decreases, the overall metrics remain consistently high, demonstrating the strong generalization capability and robustness of ScanTalk.
\begin{wraptable}[8]{r}{0.45\linewidth}
\vspace{-0.7cm}
\caption{Distance between generated motions and true scan motions. Here, $\sigma= 0.1$ for the varifold metric.}
\label{tab:temporal_consistency_comparison}
\resizebox{\linewidth}{!}{%
\begin{tabular}{l@{\hspace{0.5cm}}ccc}
\toprule
Trained on         & HD & CD ($\times 10^{-5}$) & Varifold ($\times 10^{-5}$) \\
\midrule
Reg (CH) & 0.038 & 7.60 & 1.74\\
Rmsh (CH) & 0.042 & 7.89 & 1.98 \\
Rmsh+H (CH) & 0.044 & 7.92 & 2.02 \\
Rmsh+H+GN (CH) & 0.039 & 7.98 & 2.03 \\
\bottomrule
\end{tabular}}
\label{tab:inference_on_scans}
\end{wraptable}
Finally, we evaluate our model performance on raw scan meshes with these different training strategies. We animate the first frame of raw scan sequences from VOCA. We compare the predicted motions against ground truth raw scans sequences. To do this, we use three other unregistered metrics: the mean Chamfer distance, the Hausdorff distance and the varifold distance introduced in Section~\ref{subsubsec:unregistered_setting_method}. As shown in Table~\ref{tab:inference_on_scans}, training on either registered or unregistered datasets with unregistered losses yields very similar results when tested on raw scans. This further highlights the robustness of ScanTalk across both registered and unregistered settings.

\section{Discussions and Conclusions}
By accommodating any 3D face mesh topology, ScanTalk offers versatility and robustness, making it applicable across a diverse range of datasets and scenarios. This flexibility not only resolves the constraints imposed by fixed-topology models but also simplifies the process of integrating new and varied mesh structures into the animation pipeline. 
Furthermore, the introduction of novel training schemes and loss functions that enable effective handling of unregistered meshes marks a significant step forward from previous methods that required mesh correspondence. This improvement facilitates the practical use of ScanTalk in real-world applications, where mesh registration is often impractical. 

\noindent Additionally, our work highlights the need for more comprehensive evaluation metrics in the field of 3D talking heads. The limitations of standard metrics, such as Lip-Vertex Error (LVE) and Mean Vertex Error (MVE), have been identified, and we propose new, more accurate metrics to address these shortcomings. This evaluation framework provides a clearer, more thorough assessment of animation quality and lip-sync accuracy.
Overall, ScanTalk's contributions are helpful for future research and applications in speech-driven facial animation, setting new benchmarks for flexibility, evaluation, and practical utility. Our findings underscore the potential of ScanTalk to improve the generation and assessment of 3D facial animations, with implications extending to fields such as film production, gaming, virtual reality, and beyond. 

%\section{Ethical Considerations}
\noindent We acknowledge the ethical implications associated with generating 3D facial animations. The creation of synthetic content using 3D faces carries inherent risks, including the potential for both intentional misuse and unintended societal consequences. We underscore the importance of a human-centered approach in the development and deployment of such technologies.
A human-centered design philosophy is critical for ensuring that technology serves and benefits people. The aim of our work is to advance research by addressing an open problem in the literature, with the broader goal of supporting meaningful and beneficial applications. While speech-driven facial animation has many potential uses, some positive, others potentially harmful, we stress the importance of responsible usage and rely on end users to apply this technology ethically and appropriately.
\vspace{-0.2cm}
\section*{Acknowledgments}
This work is supported by the    \href{https://geogen3dhuman.univ-lille.fr}{CNRS Int. Research Project GeoGen3DHuman}. This work was also partially supported by ``Partenariato FAIR (Future Artificial Intelligence Research) - PE00000013, CUP J33C22002830006" funded by NextGenerationEU through the italian MUR within NRRP, project DL-MIG. This work was also partially funded by the ministerial decree n.352 of the 9th April 2022 NextGenerationEU through the italian MUR within NRRP. This work was also partially supported by the project 4DSHAPE ANR-24-CE23-5907 of the French National Research Agency (ANR), and by Fédération de Recherche Mathématique des Hauts-de-France (FMHF, FR2037 du CNRS). 
This work was also partially supported by the AI4Debunk project (HORIZON-CL4-2023-HUMAN-01-CNECT grant n.101135757).
\vspace{-0.2cm}

% \begin{appendices}

% \section{Section title of first appendix}\label{secA1}

% An appendix contains supplementary information that is not an essential part of the text itself but which may be helpful in providing a more comprehensive understanding of the research problem or it is information that is too cumbersome to be included in the body of the paper.

% %%=============================================%%
% %% For submissions to Nature Portfolio Journals %%
% %% please use the heading ``Extended Data''.   %%
% %%=============================================%%

% %%=============================================================%%
% %% Sample for another appendix section			       %%
% %%=============================================================%%

% %% \section{Example of another appendix section}\label{secA2}%
% %% Appendices may be used for helpful, supporting or essential material that would otherwise 
% %% clutter, break up or be distracting to the text. Appendices can consist of sections, figures, 
% %% tables and equations etc.

% \end{appendices}

\bibliographystyle{plainnat}
\bibliography{bibliography}

\end{document}